\documentclass[letterpaper]{article} 
\usepackage{aaai24}  
\usepackage{times}  
\usepackage{helvet}  
\usepackage{courier}  
\usepackage[hyphens]{url}  
\usepackage{graphicx} 
\usepackage{natbib}  
\usepackage{caption} 
\usepackage{algorithm}
\usepackage{algorithmic}

\usepackage{newfloat}
\usepackage{listings}
\DeclareCaptionStyle{ruled}{labelfont=normalfont,labelsep=colon,strut=off} 
\floatstyle{ruled}
\newfloat{listing}{tb}{lst}{}
\floatname{listing}{Listing}
\usepackage{amsfonts, amsmath, amssymb}
\usepackage{IEEEtrantools}

\newcommand{\mcal}[1]{\mathcal{#1}}
\newcommand{\PP}{\mathbb P}
\newcommand{\EE}{{\mathbb{E}}}
\def\Real{\mathbb{R}}
\usepackage{dsfont} 
\newcommand{\bm}{\boldsymbol}
\newcommand{\x}{{\bm{x}}}
\newcommand{\wh}{\hat}

\newcommand\independent{\protect\mathpalette{\protect\independenT}{\perp}}
\def\independenT#1#2{\mathrel{\rlap{$#1#2$}\mkern2mu{#1#2}}}

\usepackage{booktabs}
\usepackage{multirow}
\usepackage{makecell}

\usepackage{amsthm}
\newtheorem{theorem}{Theorem}[section]
\theoremstyle{plain}
\newtheorem{definition}{Definition}[section]

\newtheorem{proposition}{Proposition}[section]
\newtheorem{assumption}{Assumption}

\AtEndDocument{\refstepcounter{proposition}\label{finalthm}}

\title{Learning Fair Policies for Multi-stage Selection Problems from Observational Data}
\author{
    Zhuangzhuang Jia\textsuperscript{\rm 1},
    Grani A. Hanasusanto\textsuperscript{\rm 1},
    Phebe Vayanos\textsuperscript{\rm 2},
    Weijun Xie\textsuperscript{\rm 3}
}
\affiliations{
    \textsuperscript{\rm 1}Department of Industrial and Enterprise Systems Engineering,
University of Illinois Urbana-Champaign\\
    \textsuperscript{\rm 2}Center for Artificial Intelligence in Society, University of Southern California \\

    \textsuperscript{\rm 3}H. Milton Stewart School of Industrial and Systems Engineering, Georgia Institute of Technology \\
    zj12@illinois.edu, gah@illinois.edu, phebe.vayanos@usc.edu, wxie@gatech.edu
}

\usepackage{bibentry}

\begin{document}

\maketitle

\begin{abstract}
We consider the problem of learning fair policies for multi-stage selection problems from observational data. This problem arises in several high-stakes domains such as company hiring, loan approval, or bail decisions where outcomes (e.g., career success, loan repayment, recidivism) are only observed for those selected. We propose a multi-stage framework that can be augmented with various fairness constraints, such as demographic parity or equal opportunity. This problem is a highly intractable infinite chance-constrained program involving the unknown joint distribution of covariates and outcomes.  Motivated by the potential impact of selection decisions on people’s lives and livelihoods, we propose to focus on interpretable linear selection rules. Leveraging tools from causal inference and sample average approximation, we obtain an asymptotically consistent solution to this selection problem by solving a mixed binary conic optimization problem, which can be solved using standard off-the-shelf solvers. We conduct extensive computational experiments on a variety of datasets adapted from the UCI repository on which we show that our proposed approaches can achieve an 11.6\% improvement in precision and a 38\% reduction in the measure of unfairness compared to the existing selection policy.  
\end{abstract}

\section{Introduction}

Selection problems are very common decision-making problems in many high-stakes domains, such as company hiring, college admission, and loan audit. Given a set of candidates, a decision-maker aims to select a fixed fraction of them with objectives such as hiring the most talented candidates, admitting the most qualified students, or selecting the applicants who are most likely to repay the loan. Often, selection problems are under a multi-stage setup. In hiring, for example, candidates are initially chosen for interviews based on their r\'{e}sum\'{e}s and the final selection is subsequently made from those who have been interviewed. 

Substantial evidence points to the existence of discrimination in many selection problems that involve prejudiced outcomes for individuals or groups based on \textit{sensitive} attributes like gender, race, ethnicity, nationality, disability status, or religion. For example, job applicants with African-American names are found to receive far fewer callbacks for each r\'{e}sum\'{e} they send out~\cite{bertrand2004emily}. Also, in the Canadian labor market, there exists notable discrimination towards applicants with foreign experience or those with Indian, Pakistani, Chinese, and Greek names compared with English names~\cite{oreopoulos2011skilled}. In college admission, a recent study reveals that typical Asian American applicants would see their average admit rate rise by 19\% if treated as white applicants~\cite{arcidiacono2022asian}. Additionally, recent research finds that Black-owned businesses received loans that were approximately 50\% lower than observationally similar White-owned businesses through the Paycheck Protection Program during Covid-19~\cite{atkins2022discrimination}. 

With the growing availability of data and the empirical success of machine learning, data-driven decision-making is increasingly being used in many selection problems~\cite{li2021algorithmic,ahmad2022academic,marques2022delivering}. As a result, much recent research focuses on promoting fairness and mitigating discrimination toward candidates from certain groups~\cite{barocas2017fairness,green2019principles,aghaei2019learning}. One prevalent approach to addressing fairness concerns during the training process is either by integrating fairness constraints into the training process~\cite{zafar2017fairness, zafar2019fairness,wang2021wasserstein,jo2022learning}, or penalizing discrimination using the fairness-driven regularization terms~\cite{kamishima2012fairness,berk2017convex,ye2020unbiased}. 

Often, the models are trained and evaluated on historical datasets containing fully observed outcomes and covariates. However, this raises questions about the possible harm of the deployed models as the training data may reflect implicit biases by humans who may unconsciously be in favor of certain groups of people~\cite{greenwald1995implicit,barocas2016big}. For example, in the hiring setup, the available dataset only contains full covariates of people who got hired, and the outcome measure is whether they are ``qualified" or ``not qualified"; on the other hand, we do not have access to the outcomes of candidates who were not hired. One may use a trained model based on the candidates with full covariates and outcomes to select candidates. However, when deployed to assess the qualification of future candidates, the model may exhibit significant unfairness, even if it satisfies the fairness constraints during the training process~\cite{kallus2018residual}. This is because, in the real world, the candidates have diverse profiles, unlike the training dataset that only contains profiles of hired candidates.

This paper considers the problem of learning fair policies for multi-stage selection problems in socially sensitive domains (e.g., employment, education, finance) given a labeled observational dataset containing one (or more) protected attribute(s). The main desiderata for such a framework are: \emph{(1)} Maximizing precision: the decision-maker wants to maximize his/her utility by hiring/admitting/approving as many ``qualified'' candidates among those selected as possible; \emph{(2)} Possible to augment with arbitrary fairness notions: in different socially sensitive settings, the decision-maker needs to consider the proper legal, ethical, and social standards in choosing the appropriate fairness measure; \emph{(3)} Applicable to the (potentially) biased real-world data: in the presence of a selection bias in the observational data, our model must be able to learn the fair policy for future candidates instead of those ``recorded'' candidates in the observational data. Next, we summarize the state-of-the-art in related work and highlight the need for a unifying framework that addresses these desiderata.

\subsection{Related Work}

\paragraph{Selection Problems.} \citet{kleinberg2018selection} study selection problem with implicit bias and analyze the Rooney Rule in the selection process. They show that this rule can not only improve the representation of the disadvantaged group but also lead to higher payoffs for the decision-maker. \citet{celis2020interventions} investigate the ranking problem (where the selection problem can be seen as a special case) under implicit bias and obtain similar results. \citet{khalili2021improving} study the possibility of using the exponential mechanism to address both privacy and unfairness issues. They show that this mechanism can be used as a post-processing step to improve the fairness and privacy of the pre-trained model. All these works focus on one-shot decision processes, whereas our proposed framework is applicable to multi-stage selection processes. 

\citet{emelianov2019price} study an optimal multi-stage selection problem and propose a simple model based on a probabilistic formulation. Their model, however, assumes perfect statistical knowledge of the joint distribution of covariates and outcome labels without bias. Moreover, their policies are not consistent, as they ignore the issue that the selection probability at a stage depends on which candidates were selected in the previous stages. \citet{khalili2021fair} consider a selection problem where sequentially arriving applicants apply for a limited number of positions/jobs, and the decision-maker accepts or rejects the given applicant using a pre-trained supervised learning model at each time step. Unlike their model, we consider the setting where additional covariates can only be revealed at later stages for the subset of selected individuals, whereas they assume all covariates are observable for each applicant.

\paragraph{Mixed Integer Programming (MIP).} There is a growing interest in using MIP to address machine learning tasks~\cite{bertsimas2017optimal,taskesen2020distributionally,maragno2021mixed,aghaei2021strong,jo2022learning}. \citet{aghaei2019learning} introduce a versatile MIP framework for learning optimal and fair decision trees. They show that their proposed framework yields non-discriminative decisions at a lower price to overall accuracy. \citet{ye2020unbiased} study fair classification problems and propose a framework that can be recast as mixed-integer convex programs. \citet{wang2021wasserstein} propose a distributionally robust classification model with a fairness constraint that encourages the classifier to be fair in view of the equality of opportunity criterion. They reformulate the model as a mixed binary conic optimization problem that can be solved using off-the-shelf solvers. Note that all of the above works focus on classification problems under a one-stage setup, whereas our work considers the general multi-stage selection problems.

\paragraph{Inverse Probability Weighting (IPW).} IPW is a common method to reduce selection bias and has been used in several fairness-related works. \citet{kallus2018residual} study a similar setting of the censored dataset and characterize the problem of residual unfairness. They show how to use IPW to estimate and adjust fairness metrics. However, they only focus on the one-stage static classification setup. \citet{nabi2018fair} consider the problem of fair statistical inference involving outcome variables and use the IPW method to estimate the natural direct effect. \citet{khademi2019fairness} study the problem of detecting group unfairness. They introduce fair on average causal effect -- a definition of group fairness grounded in causality and show how to use IPW to estimate fair on average causal effect and use the resulting estimates to detect and quantify discrimination based on specific attributes. \citet{kilbertus2020fair} analyze consequential decision-making using imperfect predictive models. They use IPW to compute the expected overall profit of a given policy. To the best of our knowledge, IPW has not been used to deal with our multi-stage selection problem.

\paragraph{Biased Data.} There are many works on the interplay between biased data and fairness in classification~\citep{blum2019recovering,kilbertus2020fair,rezaei2021robust,jo2021learning,liao2023social}. \citet{lakkaraju2017selective} study the ``selective labels'' problem. They develop an approach that harnesses the heterogeneity of human decision-makers. Specifically, the paper assumes that the decision-makers differ in the thresholds they use for
their yes-no decisions, but the paper does not consider the fairness of the learned policy. \citet{goel2021importance} established a causal framework to analyze the effect of missing data on the fairness of downstream tasks. The authors consider a multi-stage decision-making process and propose a decentralized approach, which is different from ours. Also, we do not assume the availability of the outcome label selected candidate at every stage.

\subsection{Proposed Approach and Contributions}
Our main contributions are summarized as follows:
\begin{enumerate}
    \item We propose a framework for learning fair policies in multi-stage selection problems from observational data. Our framework can be augmented with various fairness constraints, such as demographic parity or equal opportunity. 
    \item Leveraging tools from causal inference and sample average approximation, we obtain an asymptotically consistent solution to this selection problem by solving a mixed binary conic optimization problem using standard off-the-shelf solvers.
    \item We conduct experiments on synthetic and real-world datasets. The superiority of our model is observed through substantial precision improvement and unfairness reduction compared to the existing selection policy.
\end{enumerate}

\paragraph{Notations.} Vectors are printed in bold letters, while scalars are printed in regular font. For any $t \in \mathbb N$, we define $[t]$ as the index set $\{1,\dots,t\}$. We denote by $\mathbf{e}$ as the vector of all ones whose dimension will be clear from the context. For any set $\mcal S$, we use $| \mcal S|$ to denote its cardinality. For any logical expression $\mcal E$, the indicator function $\mathds{1}(\mcal E)$ admits value 1 if $\mcal E$ is true and 0 otherwise. We denote by $\delta_{\bm \xi}$ the Dirac distribution concentrating unit mass at $\bm \xi \in \Xi$ where $\Xi$ is the support set of the distribution. We use $\Real_{+}$ to denote the set of nonnegative real numbers and $\Real_{++}$ to denote the set of strictly positive real numbers. 
 
\section{Multi-stage Selection Problem}
We formalize the multi-stage selection problem in this part. All proofs are included in the supplemental Appendix A.

\subsection{Problem Setup}
\label{sec:multi-problem}

We consider the multi-stage problem of learning a fair selection policy with~$T$ stages. Assume that there are~$n$ candidates from two demographic groups, distinguished based on a single sensitive attribute $A \in \mathcal{A} = \{0,1\}$ that represents their group membership. This sensitive attribute could be information such as gender, race, or age group, which differentiates privileged and unprivileged individuals. 

At stage $t \in [T-1]$, for those $n_{t-1} $ candidates that passed stage $t-1$, the decision maker observes an extra covariate vector $\bm X^t \in \mathcal{X}^t \subseteq \mathbb{R}^{d_t}$ that is not available in the previous stages. In the real world, $\bm X^t$ can represent predictors delineating qualifications, creditworthiness, criminal history, etc. Next, based on all available features $\bm X^{[t]}:= (\bm X^1,\dots, \bm X^t) \in \mathbb{R}^{d_1+\dots+d_t}$, $n_t \leq \overline{\alpha}_t n$ candidates are selected to advance to the next stage where $0 <\overline{\alpha}_{t} \leq 1$ denotes the predefined upper bound selection ratio by the decision-maker. This process continues until the final stage~$T$, where all covariate vectors $\bm X^{[T]}$ of those who were selected at stage~$T-1$ are revealed. The decision maker then selects $\underline{\alpha}_Tn \leq n_T \leq \overline{\alpha}_Tn$ candidates from those available at stage~$T$, where $0 < \underline{\alpha}_T \leq \overline{\alpha}_T \leq \dots \leq \overline{\alpha}_1 \leq 1$. Unlike the previous stages, the final stage has a lower bound selection ratio $\underline{\alpha}_T$. In the real world, $\underline{\alpha}_T$ represents the minimum hiring/admission rate by the decision-maker. For example, the university has a minimum admission rate -- a level at which the school may lose money on tuition, federal aid, or not using resources like faculty and classrooms to capacity.

Additionally, we assume each candidate has a binary outcome label $Y \in \mathcal{Y} = \{0,1\}$, which can only be observed if the candidate were selected to be hired (i.e., made it to the last stage). Without loss of generality, we use the positive response to indicate a positive (good) outcome, such as the candidate is qualified, the applicant repays the loan, or the individual does not recidivate. 

\begin{figure}[htb!]
\centering
\includegraphics[scale=0.12]{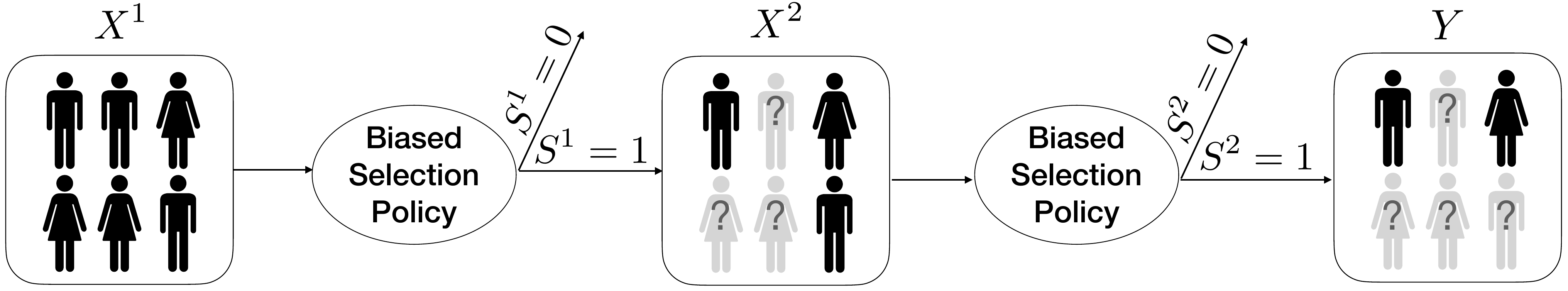} 
\caption{Two-stage selection problem: $\bm X^2$ is observable only for those selected in the first stage ($S^1=1$). Outcome label $Y$ is only observable for those selected in both stages ($S^1=S^2=1$).}
\label{fig:multi-data}
\end{figure}
Assume that we possess a training dataset containing $N$ samples of the form $\{\hat \x^{[T]}_i, \hat a_i, \hat y_i \}_{i=1}^N$ that are generated from an unknown joint probability distribution $\PP$ of the random vector $(\bm X^{[T]}, A, Y)$. We denote by $I^t$ the set of selected candidates at stage $t-1$. In other words, $I^t$ includes all candidates with covariates up to time $t$. Thus, we have $[N] = I^0 \supseteq I^1 \supseteq I^2 \supseteq \dots \supseteq I^T$. The set $I^T$ collects the indices of those candidates whose entire covariates $\bm X^{[T]}$ and outcome label $Y$ are observable. Figure~\ref{fig:multi-data} shows the biased training dataset of a two-stage selection process.

Let $S^t \in \{0, 1\}$ be the selected outcomes where $S^t = 1$ if the candidate is selected at stage $t$. In this paper, we make the following assumptions. 
\begin{assumption}[Conditional Exchangeability]\label{assump:3}
Whether a candidate is selected or not at stage $t \in [T]$ is independent of all future covariates $(\bm X^{t+1}, \dots, \bm X^T, Y)$, and mathematically,
 \begin{align*}
     (\bm X^{t+1}, \dots, \bm X^T, Y) \independent S^t ~|~ \bm X^{[t]}, S^{[t-1]} & = 1 \\
     & \forall \bm X^{[t]} \in \mcal X^{[t]}. 
 \end{align*}
\end{assumption}
Assumption~\ref{assump:3} implies that at each stage $t$, whether a candidate is selected or not depends only on the available covariates $\bm X^{[t]}$ and that there are no unmeasured confounders that affect both $(\bm X^{t+1}, \dots, \bm X^T, Y)$ and the selection decision $S^t$. In reality, the selection decision at each stage $t$ is only based on observed covariates up to that stage. Hence, we can infer the outcome distribution for individuals who were not selected in the observational data by looking at their counterparts with the same (or similar) $\bm X^{[T]}$ values who got selected. 

\begin{assumption}[Positivity]\label{assump:4}
At stage $t \in [T]$, the probability of being selected is strictly positive for any candidate's covariate values, i.e.,
 \begin{align*}
    \PP (S^t=1 | \bm X^{[t]}) >0 \quad \forall \bm X^{[t]} \in \mcal X^{[t]}.
 \end{align*}
\end{assumption}
The positivity assumption states that any candidate should have a positive probability of being selected at any stage. Otherwise, there is no information about the distribution of the outcomes for some covariates, and we will not be able to make inferences about it.

In the multi-stage fair selection problem, at each stage $t$, in view of the candidate’s information $\bm X^{[t]}$, the decision-maker aims to find a policy $\mcal C_t: \mcal X^1 \times \dots \times \mcal X^t \to \mcal Y$ that determines whether the candidate proceeds to the next stage or not. In the real world, those finally selected candidates can represent hired, admitted, or approved candidates. The decision-maker wants to maximize his/her utility by hiring/admitting/approving more ``qualified" candidates among those selected. Hence, we use the precision $\PP ( Y = 1~|~ \mcal C_T (\bm X^{[T]}) = 1 )$ as our performance metric. 

\paragraph{Fairness Notions} 
In the machine learning literature, different notions of fairness can generally be classified into \textit{individual fairness} and \textit{group fairness}. Both perspectives have their advantages and limitations. In this paper, we concentrate on the \textit{group fairness} due to its straightforward definition and comprehensibility for decision-makers. Furthermore, in practice, many people prioritize assessing and enforcing \textit{group fairness}~\cite{los2018report}. 

We now briefly explain several commonly used group fairness notions based on the sensitive attribute $A$ and introduce our unfairness measure. \textit{Demographic Parity} requires the probability of being selected to be equal across different demographic groups~\citep{calders2009building}. \textit{Equal Opportunity} requires the true positive rate (TPR) to be equal across different demographic groups~\citep{hardt2016equality}. \textit{Conditional Statistical Parity} requires the probability of being selected to be equal across different demographic groups, conditional on some legitimate covariate(s) indicative of risk~\citep{corbett2017algorithmic}. Additional fairness concepts include disparate impact~\cite{feldman2015certifying} and disparate mistreatment~\cite{zafar2017fairness} criteria. We refer the interested readers to~\citet{pleiss2017fairness,chouldechova2020snapshot,mehrabi2021survey} for extensive reviews of the literature.

In the real world, the decision-maker needs to 
consider the proper legal, ethical, and social context in choosing the proper fairness measure. For example, \textit{Demographic Parity} could be chosen to address representation disparities, which can be important for promoting diversity. And the decision-maker may choose \textit{Equal Opportunity} to ensure fairness among those ``qualified" candidates. For a given fairness notion, we define the unfairness measure as follows.
\begin{definition}[Unfairness measure]\label{def:unfair_measure_mu}
For a given policy $\mcal C_T $, the unfairness measure is defined as the absolute disparity of the respective statistical metric across groups, and we denote it using $\mathds U(\mcal C_T(\bm X^{[T]}), \PP)$.
\end{definition}
Here, we focus on fairness in the final stage. However, it is worth noting that fairness can also be enforced at every stage by employing similar unfairness measures. The larger the value of $\mathds U(\mcal C_T(\bm X^{[T]}), \PP)$, the more unfair our selection policy is. In other words, it measures how biased the selection policy is across the privileged and unprivileged groups. Due to the page limit, we will concentrate on unfairness measures using the \textit{Equal Opportunity} notion, which is defined as follows:
\begin{align*}\label{eq:unfmeasure}
    \mathds U(\mcal C_T(\bm X^{[T]}), \PP) = | \PP ( \mcal C_T(\bm X^{[T]}) = 1 \;\vert\; A = 1, Y = 1 ) \\
     - \PP ( \mcal C_T(\bm X^{[T]}) = 1 \;\vert\; A = 0, Y = 1 )|.
\end{align*}
Nonetheless, our approach is flexible enough to accommodate other notions of fairness.

\begin{figure}[htb!]
\centering
\includegraphics[scale=0.21]{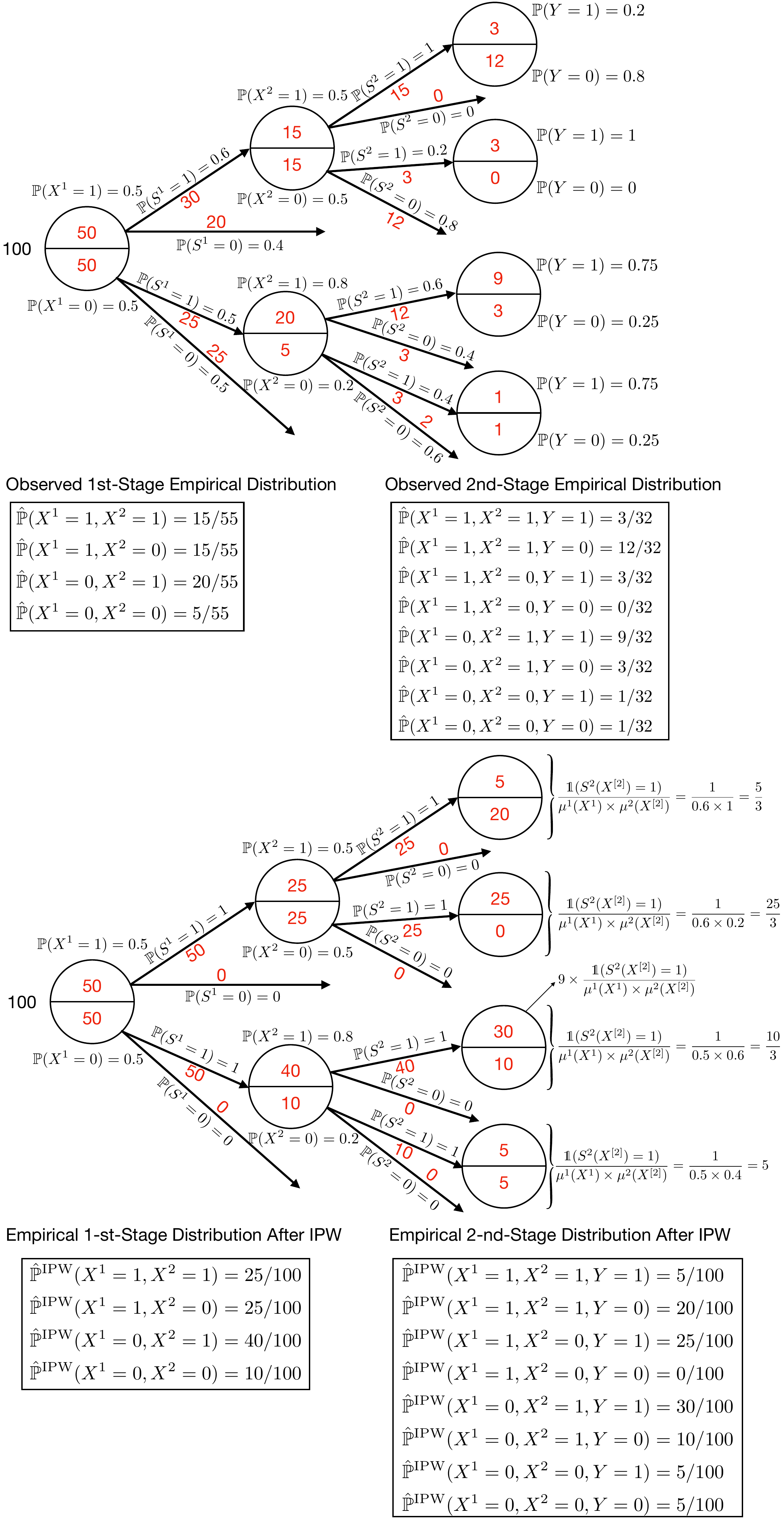} 
\caption{Evaluation of the performance of a counterfactual two-stage selection process (shown in the bottom tree) using data collected by an observed selection process (shown in the top tree) using the IPW estimator. The dataset consists of 100 candidates with binary features in each stage.}
\label{fig:ipw_multi}
\end{figure}
\subsection{Mathematical Formulation}
\label{sec:math_formulation}
Based on the previous problem description, we now present our proposed infinite chance-constrained program for learning optimal multi-stage fair selection policy, as follows:
\begin{equation} \label{eq:generalmodel_multi}
\renewcommand*{\arraystretch}{1.2}
    \begin{array}{cl}
    \max\limits_{\{ \mcal C_t(\cdot)\}_{t=1}^T}& \PP ( Y = 1~|~ \mcal C_T ( \bm X^{[T]})=1  ) \\
    \textrm{s.t.} & \PP ( \mcal C_t(\bm X^{[t]}) = 1 ) \leq \overline{\alpha}_t \hspace{.23in} \forall t \in [T]\\ 
     & \PP ( \mcal C_T(\bm X^{[T]}) = 1 ) \geq \underline{\alpha}_T \\ 
    &  \mcal C_{t+1}(\bm X^{[t+1]}) \leq \mcal C_{t}(\bm X^{[t]}) \\
    &\qquad  \forall (\bm X^{[t]}, \bm X^{[t+1]})\;\; \forall t \in [T-1]\\ 
    & \mathds{U} (\mcal C_T(\bm X^{[T]}), \PP) \le \eta.
    \end{array}    
\end{equation} 
The objective function aims to maximize the ratio of ``qualified" candidates to those who advance to the final stage, e.g., hired candidates. The first two constraints represent our selection ratio requirements. The penultimate constraint ensures the decisions are consistent -- that is, candidates that were dropped at stage $t$ can no longer be selected at the next stage $t+1$.  The last constraint corresponds to the fairness constraint, where $ \eta\in[0,1]$ represents the unfairness tolerance of the decision-maker. 

Problem~\eqref{eq:generalmodel_multi} is challenging because \emph{(i)} it optimizes over functions; \emph{(ii)} due to selection bias, we cannot observe outcome labels $Y$ for those who did not get selected in the training data; \emph{(iii)} we do not know the true distribution $\PP$ of $(\bm X^{[T]}, A, Y)$, and even if $\PP$ were known, the probabilistic program~\eqref{eq:generalmodel_multi} is computationally difficult since the problem of computing the probability of an event involving multiple random variables belongs to the complexity class \#P-hard~\cite{dyer1988complexity}. 

To address the first challenge and to also make the decisions transparent and hence accountable, we focus on the interpretable linear selection policies $\mcal C_t(\bm X^{[t]})$ parameterized by slope parameters $\bm W^{[t]} = (\bm w_t^1, \dots, \bm w_t^t) \in \mathbb{R}^{d_1+\dots+d_t}$ and an offset $b_t \in \mathbb{R}$. The selection decision is then determined through an indicator function of the form $\mcal C_t(\bm X^{[t]}) = \mathds{1}(\bm W^{[t]} \cdot \bm X^{[t]} + b_t > 0)$, where $\bm W^{[t]} \cdot \bm X^{[t]} = \sum_{j=1}^t{\bm w_t^j}^\intercal \bm X^j$.

For the second challenge, one may propose to include only the selected candidates in the training dataset without any modification; however, the resulting dataset is not i.i.d. due to selection bias. To tackle this challenge, we employ the IPW scheme to evaluate the performance of a \emph{counterfactual} selection policy. Specifically, we assume that the historical selection in the data follows a \emph{logging policy} $\{ \bm \mu^t \}_{t=1}^T$. For a candidate with covariates $\x^{[t]} = (\x^1, \dots, \x^t)$, we have $\bm \mu^t(\x^{[t]}):= \PP(S^t= 1| \bm X^{[t]}=\x^{[t]}, S^{[t-1]} =1 )$, i.e., it represents the selection probability of a candidate with covariate $\x^{[t]}$ at stage $t$. \citet{horvitz1952generalization} originally proposed IPW as a method to estimate causal quantities such as expected values of counterfactual outcomes, average treatment effects, and risk ratios. IPW involves reweighting the outcome of each selected candidate $i \in I^T$ by the inverse of their \emph{propensity score}, denoted as $\bm \mu^t(\x_i^{[t]})$. This reweighting creates a \emph{pseudo-population} where all candidates in the data are hypothetically selected. This allows for the estimation of the distribution of unobserved counterfactual selected outcomes for all candidates. We then estimate a counterfactual selection policy by reweighting each selected individual~$i \in I^T$ at stage $t$ by $\beta_i^t = {\mathds{1}(S^t_i = 1)}/{\prod_{j=1}^t \hat {\bm \mu}^j(\x_i^{[j]})}$ as illustrated in Figure~\ref{fig:ipw_multi}; see also \cite{bottou2013counterfactual}. Here, $\hat {\bm \mu}^t$ is an estimator of $\bm \mu^t$ that can be obtained using machine learning techniques like logistic regression by fitting a model to $\{ \hat \x_i^{[t]}, S_i^t \}_{i \in I^t}$.  

Finally, to tackle the last challenge, we use sample average approximation to approximate the true distribution empirically. In a data-driven setting, at stage $t$, we only have access to $|I^t|$ training samples generated from $\PP$, and we define $\hat \PP^{\text{IPW}}$ to be the empirical distribution supported on $\{\hat \x^{[t]}_i,\hat a_i, \hat y_i \}_{i\in I^T}$ after applying IPW: 
\begin{align*}
    \hat \PP^{\text{IPW}} = \sum_{i=1}^{|I^t|} \frac{ \beta_i^t }{\mathbf{e}^\intercal \bm \beta^t} \delta_{(\hat \x_i^{[t]}, \hat a_i, \hat y_i) }.
\end{align*} 

\subsection{MIP Reformulation}
\label{sec:mip_formulation}
Using the aforementioned methods, we obtain the following finite-dimensional chance-constrained program:
\begin{equation} \label{eq:exactmodel_multi}
\renewcommand*{\arraystretch}{1.2}
    \begin{array}{cl}
    \max & \hat{\PP}^{\text{IPW}} ( Y = 1~|~ \mcal C_T (\bm X^{[T]})=1  ) \\
    \mathrm{s.t.} & \bm W^{[t]} \in \mathbb{R}^{d_1+\dots+d_t}, ~ b_t \in \Real \hspace{.42in} \forall t \in [T]   \\
    & \hat{\PP}^{\text{IPW}} (\bm W^{[t]} \cdot \bm X^{[t]} + b_t > 0 ) \leq \overline{\alpha}_t \;\; \forall t \in [T] \\
    &  \hat{\PP}^{\text{IPW}} ( \bm W^{[T]} \cdot \bm X^{[T]} + b_T > 0 ) \geq \underline{\alpha}_T \\ 
    &  \mcal C_{t+1}( \bm X^{[t+1]}) \leq \mcal C_{t}( \bm X^{[t]})\\ 
    &\hspace{.75in} \forall (\bm X^{[t]}, \bm X^{[t+1]}) \;\; \forall t \in [T-1]\\ 
    & \mathds{U} (\mcal C_T( \bm X^{[T]}), \hat{\PP}^{\text{IPW}}) \le \eta.
    \end{array} 
\end{equation} 
Program~\eqref{eq:exactmodel_multi} allows decision-makers to explicitly bound the unfairness measure in the training set using $\eta$. Unfortunately, it remains challenging to transform~\eqref{eq:exactmodel_multi} into an exact mixed-integer conic representable formulation that can be solved using standard MIP solvers. To see this, consider the lower bound selection ratio constraint at the final stage, which can be further represented as 
\begin{equation*}
     \sum_{i=1}^{|I^T|} \beta_i^T \mathds{1}( \bm W^{[T]}\cdot \bm X^{[T]} + b_T \leq 0 ) \leq (1-\underline{\alpha}_T)  \mathbf{e}^\intercal \bm \beta^T.
\end{equation*}
It can be verified that for $\underline{\alpha}_T \in (0,1)$, the feasible region of $(\bm W^{[T]}, b_T)$ with such constraint is an open set that cannot be exactly reformulated as a bounded MIP problem~\cite{jeroslow1987representability}. If $1-\underline{\alpha}_T < 1/(\mathbf{e}^\intercal \bm \beta^T )$, then $(\bm W^{[T]}, b_T)$ must satisfy $\bm W^{[T]} \cdot \bm X^{[T]} + b_T > 0 \; \forall i \in I^T$.

To address this issue, we propose a conservative approximation to~\eqref{eq:exactmodel_multi}. Firstly, for the selection ratio constraint, we change the inequality sign of the function $\mathds{1}( \bm W^{[T]}\cdot \bm X^{[T]} + b_T \leq 0)$ to a \textit{strict} inequality. Furthermore, to ensure robustness, we modify the right-hand side to a positive quantity $\epsilon$ and use an inner approximation. Next, for the fairness constraint, leveraging the finite cardinality of $\mcal A$ and $\mcal Y$, we can decompose $\hat \PP^{\text{IPW}}$ using its conditional measures $\hat \PP^{\text{IPW}}_{ay} (\cdot) = \hat \PP^{\text{IPW}}( \cdot ~|~A = a, Y=y)$. We now define the $\epsilon$-unfairness measure $\mathds{U}_{\epsilon}$ as 
\begin{align*} 
    \max \left\{ \begin{array}{l}
    \hat \PP^{\text{IPW}}_{01} (\bm W^{[T]} \cdot \bm X^{[T]} + b_T > 0 )  \\
     \quad - \hat \PP^{\text{IPW}}_{11} ( \bm W^{[T]} \cdot \bm X^{[T]} + b_T \geq \epsilon ),\\
    \hat \PP^{\text{IPW}}_{11} ( \bm W^{[T]}\cdot  \bm X^{[T]} + b_T > 0 ) \\
     \quad - \hat \PP^{\text{IPW}}_{01} (\bm W^{[T]}\cdot \bm X^{[T]} + b_T\geq \epsilon ) 
    \end{array}
    \right\},
\end{align*}
which is parameterized by a strictly positive value $\epsilon \in \Real_{++}$. Lastly, for the penultimate constraint in \eqref{eq:exactmodel_multi}, we introduce
\begin{align*}
    C_{t}^\epsilon(\bm X^{[t]}) = \mathds{1}(  \bm W^{[t]} \cdot \bm X^{[t]}+ b_t \geq \epsilon ),
\end{align*}
where $\epsilon \in \Real_{++}$. This approximation yields our proposed $\epsilon$-IPW multi-stage fair selection ($\epsilon$-IPWMFS) model:
\begin{equation} \label{eq:appromodel_multi}
\renewcommand*{\arraystretch}{1.2}
    \begin{array}{cl}
    \max & \hat{\PP}^{\text{IPW}}( Y = 1~|~ \mcal C_T (\bm X^{[T]})=1  ) \\
    \mathrm{s.t.} & \bm W^{[t]} \in \mathbb{R}^{d_1+\dots+d_t},~ b_t \in \Real \hspace{.42in}\forall t \in [T]   \\
    & \hat{\PP}^{\text{IPW}} ( \bm W^{[t]} \cdot \bm X^{[t]} + b_t > 0 ) \leq \overline{\alpha}_t \;\; \forall t \in [T] \\
    & \hat{\PP}^{\text{IPW}} (\bm W^{[T]} \cdot \bm X^{[T]} + b_T < \epsilon ) \leq 1 - \underline{\alpha}_T \\ 
    &  \mcal C_{t+1}(\bm X^{[t+1]}) \leq \mcal C_{t}^\epsilon(\bm X^{[t]}) \\ 
    &\hspace{.75in}  \forall (\bm X^{[t]}, \bm X^{[t+1]})\;\; \forall t \in [T-1]\\ 
    & \mathds{U}_{\epsilon}(\mcal C_T(\bm X^{[T]}), \hat{\PP}^{\text{IPW}} ) \le \eta. 
    \end{array}    
\end{equation} 
It is worth noting that when defining the $\epsilon$-IPWFS model~\eqref{eq:appromodel_multi}, we have the option to use three distinct values of $\epsilon$: one for the admission requirement constraint, one for the penultimate constraint, and another for $\mathds U_\epsilon$. However, to simplify the notation and avoid the need for excessive parameter tuning, we use a single parameter $\epsilon$. 
\begin{proposition}[Conservative approximation]\label{prop:mpm_conservative_multi}
        Let~$ \{ {\bm W^{[t]}}^\star, b_t^\star \}_{t=1}^T$ be an optimal solution to problem~\eqref{eq:appromodel_multi}. Then $ \{ {\bm W^{[t]}}^\star, b_t^\star \}_{t =1}^T$ is feasible in problem~\eqref{eq:exactmodel_multi}. Moreover, let~$f^\star$ and~$f^\star_{opt}$ be the optimal values of problems~\eqref{eq:appromodel_multi} and~\eqref{eq:exactmodel_multi}, respectively. Then, $f^\star_{opt} \ge f^\star$.
\end{proposition}
According to Proposition~\ref{prop:mpm_conservative_multi}, the optimal value of problem~\eqref{eq:appromodel_multi} provides a lower bound on the precision. We now present the main result of this section, which asserts that the problem~\eqref{eq:appromodel_multi} can be reformulated as a mixed binary conic optimization problem.
\begin{theorem}[$\epsilon$-IPWMFS reformulation] \label{thm:refor-probtrust_multi}
    The $\epsilon$-IPWMFS model~\eqref{eq:appromodel_multi} is equivalent to the mixed binary conic optimization problem
    \begin{equation}
    \label{eq:refor-probtrust_multi}
    \renewcommand*{\arraystretch}{1.2}
    \begin{array}{cll}
            \min & f \\
            \mathrm{s.t.} &  \bm W^{[t]} \in \mathbb{R}^{d_1+\dots+d_t},~ b_t \in \Real \hspace{.81in} \forall t \in [T]   \\
            & f \in \Real, ~\bm g_t \in  \{0, 1\}^{|I^T|},~\bm p_t \in \{0, 1\}^{|I^T|}\hspace{.1in}\forall t \in [T] \\
            & 1 \leq f \\
            & \displaystyle \sum_{i=1}^{|I^T|} \beta_i^T (g_{Ti})^2 \leq f  \sum_{i \in \mcal I_1 } \beta_i^T g_{Ti} \\
            & \bm g_t^\intercal \bm \beta^t   \leq \overline{\alpha}_t(\mathbf{e}^\intercal \bm \beta^t) \hspace{1.31in} \forall t \in [T] \\
            & \bm p_T^\intercal \bm \beta^T  \leq (1 - \underline{\alpha}_T)(\mathbf{e}^\intercal \bm \beta^T) \\
            & \bm g_{t+1}  + \bm p_{t} \leq \mathbf{e} \hspace{1.27in} \forall t \in [T-1] \\
            &  \displaystyle \frac{ \sum_{i \in \mathcal{I}_{a1}}  g_{Ti} \beta_i^T }{  \sum_{i \in \mcal I_{a1}} \beta_i^T }  + \frac{ \sum_{i \in \mathcal{I}_{a'1}}  p_{Ti} \beta_i^T }{ \sum_{i \in \mcal I_{a'1}} \beta_i^T }  - 1 \leq \eta \\
            & \hspace{1.4in} \forall (a,a') \in \{(0,1),(1,0) \} \\
            & \left.
            \begin{array}{l}
                -M(1-g_{ti}) \leq \bm W^{[t]} \cdot \hat \x_i^{[t]}+ b_t \leq M g_{ti}  \\
                  \epsilon  - \bm W^{[t]} \cdot \hat \x_i^{[t]} - b_t \leq M p_{ti} 
           \end{array}
        \right\} \\
        & \hspace{1.85in}  \forall t \in [T], ~\forall i \in I^T,
        \end{array}
    \end{equation}
    where $M$ is the big-M parameter, $\mcal I_1 = \{i \in I^T:  \wh y_i = 1 \}$, and $\mcal I_{a1} = \left\{ i \in I^T: \wh a_i = a, \wh y_i = 1\right\}$.
\end{theorem}
 We remark that problem \eqref{eq:refor-probtrust_multi} can be solved using any popular off-the-shelf solver, such as Gurobi, Mosek, or CPLEX.

\section{Numerical Experiments}
\label{sec:numerical}
In this section, we present the numerical experiments using both synthetic and real-world datasets. We consider the two-stage selection process to simplify the exposition. All optimization problems were implemented using Python 3.10 and solved by Gurobi 10.0.1. The experiments were run on an M1 Ultra CPU laptop with 64GB RAM.

\paragraph{Synthetic Data Experiments}
We first use a synthetic dataset to illustrate the importance of reweighting and the effectiveness of our fair optimization model. We simulate two-stage selection data with two subgroups, one being the minority (i.e., $A=0$). Both groups have the same Gaussian distribution of true qualification: $X \sim \mcal N(0,2)$. To create a selection bias, we set~$X^1 = X - 0.5  B + noise_1$, where $noise_1 \sim \mcal N(0, 0.5), B \sim Bernoulli(.2)$ if $A = 0$; and $B \sim Bernoulli(.1)$ if $A = 1$. Then, the candidates are selected for the next stage with probability $1/(1+e^{-X^1})$. This selection criterion ensures that every candidate has a non-zero probability of being selected, and a higher value of $X^1$ corresponds to a higher probability of selection.

Next, for those who enter the second stage, we set a more biased $X^2 = X - 0.5 \times  \mathds{1}(A=0) + 0.5\times \mathds{1} (A=1) + noise_2$, where $noise_2 \sim \mcal N(0, 0.25)$. Specifically, we tend to underestimate the qualifications of candidates from the minority group while overestimating those from the majority group. Then, based on both  $X^{1}$ and $X^2$, the candidates are selected with probability $1/(1+e^{-(0.7X^2+0.3X^1)})$. 

Lastly, we generate the outcome label for each selected candidate as $Y = \mathds{1}(X \geq 1)$. In other words, the label is only related to the true qualification and is not influenced by any sensitive attribute. 

Using the aforementioned procedure, we conducted simulations of over 200,000 candidates. The results indicate that such an existing selection policy has an overall precision of approximately 68.57\% and an overall unfairness score of 0.050. We will use these results as a benchmark to compare against our proposed methods.
\begin{table}[ht!]
    \centering
    \resizebox*{\columnwidth}{!}{%
\begin{tabular}{||c|c|c|c|c|c|c||}
\hline
\thead{Training \\ Size} &   & \thead{ Testing $1^{st}$-Stage \\  Violation Ration } & \thead{Testing $2^{nd}$-Stage \\  Violation Ration} & \thead{Testing Precision \\ (mean ± standard deviation)} & \thead{Testing \\ Unfairness}  & \thead{Training \\ Time}    \\ \hline \hline
\multirow{2}{*}{100} & IPW & 0\% & 80\%  & $82.92\% \pm 12.63\%$ & 0.0447 & 0.0768s  \\ 
                 & No-IPW &  20\% & 80\% & $69.79\% \pm 29.39\%$ & 0.1224  & 0.3579s \\  \hline
\multirow{2}{*}{200} & IPW & 0\% & 80\% & $90.22\% \pm 10.51\%$ & 0.0931 & 0.2004s  \\ 
                 & No-IPW &  0\% &  100\%  & $77.57\% \pm 16.76\%$ & 0.0135 & 13.9079s \\  \hline
\multirow{2}{*}{400} & IPW & 20\% & 40\% & $93.53\% \pm 5.93\%$ & 0.1772 &  26.4929s  \\ 
                 & No-IPW & 20\% & 80\% & $66.83\% \pm 32.31\%$ & 0.0674 & 30.8252s \\  \hline
\multirow{2}{*}{800} & IPW & 0\% & 40\% & $95.00\% \pm 4.56\%$ & 0.1742 & 55.0997s  \\ 
                 & No-IPW & 40\% & 60\% & $51.28\% \pm 37.70\%$ & 0.0322 & 100.4673s \\  \hline
\multirow{2}{*}{2000} & IPW & 0\% & 20\% & $94.72\% \pm 3.83\%$ & 0.1390 & 197.3959s  \\ 
                 & No-IPW & 60\% & 40\% & $16.05\% \pm 20.71\%$  & 0.0251 & 198.8476s \\   \hline
\multirow{2}{*}{3000} & IPW & 0\% & 0\% & $95.94\% \pm 2.13\%$ & 0.1840  & 240.3892s  \\ 
                 & No-IPW & 20\% & 80\%  & $41.91\% \pm 35.11\%$ & 0.0136 & 219.9468s \\  
\hline
\end{tabular}
}
    \caption{Out-of-sample testing results. Here we set the fairness parameter $\eta = 1$.}
    \label{tab:my_label}
\end{table}
To implement the IPW scheme, we use logistic regression to estimate the propensity score $\bm \beta^t$. For the No-IPW scheme, we assign an equal weight to each selected candidate data point by setting $\bm \beta^t = \mathbf{e}$. We conduct out-of-sample experiments with training dataset sizes $N=100,200,400,800,2000,3000$. The selection ratios are set to $\overline{\alpha}_1 = 0.7$, $\overline{\alpha}_2 = 0.35$ and $\underline{\alpha}_2 = 0.2$. The results of all experiments are averaged over five random trials. We set a time limit of 4 minutes for each trial (the final solution obtained with $N = 2000$ and 3000 samples may not be optimal). For each trial, we evaluate the performance using 10,000 independent testing samples. To ensure a fair comparison during the out-of-sample test, we randomly select candidates from previous stages if we select fewer than $\underline{\alpha}_2$ proportion candidates to meet the lower bound selection requirement. Similarly, if we select more than $\overline{\alpha}_1$ or $\overline{\alpha}_2$ proportion candidates, we randomly eliminate some candidates from those selected to meet the upper bound selection requirement. 

The out-of-sample statistics in Table~\ref{tab:my_label} showcase the superior performance of $\epsilon$-IPWMFS, as evidenced by its decreasing constraint violation ratio as the training size increases. In contrast, the No-IPW scheme consistently yields a 100\% violation ratio, rendering the generated policies unsuitable for implementation. Furthermore, IPW achieves higher precision as the training size increases, whereas the No-IPW scheme has low precision and a lower fairness score. This is due to the high violation ratio, so some candidates need to be randomly selected or eliminated to satisfy the constraints. 

\begin{figure}[ht!]
\centering\includegraphics[width=.4\textwidth]{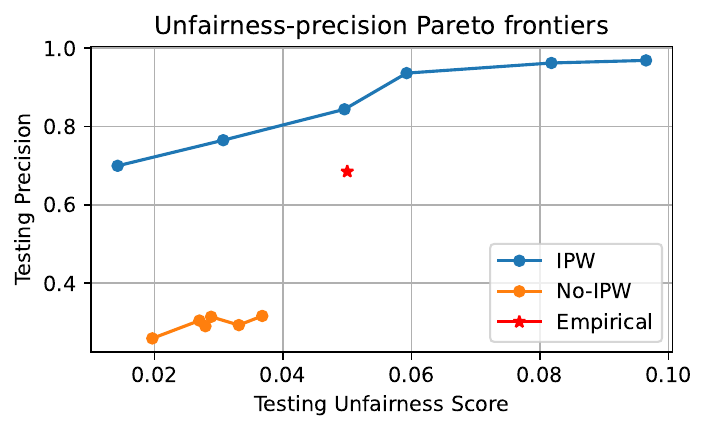}
\caption{Pareto frontiers. The red star represents the unfairness and precision of the existing selection policy.}
\label{fig:pareto}
\end{figure}

We plot the Pareto frontiers of the two schemes with training size $N=800$ in Figure~\ref{fig:pareto}. We examine models with different values of the unfairness controlling parameter $\eta$ on [0.01, 0.06] with six equidistant points, and the results were obtained from 10 independent trials. Compared to the No-IPW scheme, the IPW scheme provides higher precision for the same unfairness score. Additionally, the existing selection policy provides an unfairness score of 0.050 and a precision of 68.57\% (red star in Figure~\ref{fig:pareto}). Using our proposed $\epsilon$-IPWMFS method, the learned policy can achieve a much higher precision of 76.51\% and a lower unfairness score of 0.031. Without IPW, the resulting learned policy is not useful as the observational data cannot represent the real-world ``test'' population due to selection bias.
 
\paragraph{Experiments with Real-world Data}
In this part, we present several semi-synthetic experiments using the Adult~\cite{kohavi1996scaling}, COMPAS~\cite{angwin2022machine}, and German~\cite{dua2019uci} datasets to illustrate the effectiveness of our framework. Such semi-synthetic experiments are necessary because of the unavailability of datasets with a multi-stage selection setup. Details about the datasets are provided in the supplemental Appendix B.

For each dataset, we randomly split the covariates into two sets, one for the first-stage $\bm X^1$ and one for the second-stage $\bm X^2$. We simulate a synthetic selection process based on the following. In the first stage, we use a trained logistic regression model $f_1(\bm X^1) = P(Y = 1|\bm X^1)$ to learn the true outcome label $Y$, and a trained logistic regression model $g(\bm X^1) = P(A = 1|\bm X^1)$ to learn the sensitive attribute $A$. We create a selection bias by assigning a score to each candidate as follows: $Score_1 = 10f_1(\bm X^1)- 2B + noise_1$, where $noise_1 \sim \mcal N(0,1), B \sim Bernoulli(.2)$ if $g(\bm X^1) = 0$; and $B \sim Bernoulli(.1)$ if $g(\bm X^1) = 1$. Then the candidates are selected for the next stage with probability $1/(1+e^{-Score_1})$. This selection criterion ensures that every candidate has a non-zero probability of being selected, and a higher value of $\bm X^1$ corresponds to a higher probability of selection.

Next, for those who get into the second stage, after we see additional covariates $\bm X^2$ and sensitive attribute $A$, we use a trained logistic regression model $f_2(\bm X^{[2]})=P(Y = 1|\bm X^{[2]})$ to learn the true label $Y$. Then, we assign a score to each candidate as follows: $Score_2 = 10f_2(\bm X^{[2]})- 1.5 \times \mathds{1} (A=0)+ 1.5 \times \mathds{1} (A=1) +noise_2$, where $noise_2 \sim \mcal N(0,1)$.  The candidates are finally selected with probability $1/(1+e^{-(0.8Score_2+0.2Score_1)})$.
\begin{table}[ht]
    \centering
    \resizebox{.92\columnwidth}{!}{%
     \begin{tabular}{||p{2.5cm}|p{2.5cm} | c| c| c| c| c||} 
 \hline
 \textbf{Dataset} & \textbf{Metric} & \textbf{No-Fair} & \textbf{Fair} & \textbf{Empirical}\\
 \hline\hline
 \multirow{6}{4em}{Adult} & Precision & $\bm{0.52 \pm 0.05}$ & ${0.51 \pm 0.04~(1.92\%)}$  & 0.28 \\ 
 &Unfairness ($\mathds U$)& $ {0.26 \pm 0.19}$ & $\bm{0.17 \pm 0.13~(34.61\%)}$ & 0.22 \\
 \cline{2-5}
 & Precision & $\bm{0.45 \pm 0.05}$ & ${0.44 \pm 0.05~(2.22\%)}$ & 0.27 \\ 
 &Unfairness ($\mathds U$)& $ {0.21 \pm 0.15}$ & $\bm{0.16 \pm 0.12~(23.81\%)}$  & 0.23\\
 \cline{2-5}
 & Precision & $\bm{0.39 \pm 0.05}$ & ${0.37 \pm 0.06~(5.13\%)}$  & 0.27 \\ 
 &Unfairness ($\mathds U$)& $ {0.17 \pm 0.11}$ & $\bm{0.14 \pm 0.12~(17.65\%)}$  & 0.24\\
 \hline
 \multirow{6}{4em}{German} & Precision & $\bm{0.78 \pm 0.12}$ & ${0.78 \pm 0.13~(0\%)}$  & 0.76\\
 &Unfairness ($\mathds U$)& ${0.08 \pm 0.07}$ & $\bm{0.05 \pm 0.05~(37.5\%)}$ &  0.16 \\
 \cline{2-5}
 & Precision & $\bm{0.82 \pm 0.11}$ & ${0.76 \pm 0.13~(7.32\%)}$  & 0.76\\ 
 &Unfairness ($\mathds U$)& $ {0.11 \pm 0.09}$ & $\bm{0.05 \pm 0.03~(54.55\%)}$ & 0.17 \\
 \cline{2-5}
 & Precision & $\bm{0.73 \pm 0.09}$ & ${0.72 \pm 0.08~(1.37\%)}$  & 0.72\\ 
 &Unfairness ($\mathds U$)& $ {0.12 \pm 0.06}$ & $\bm{0.09 \pm 0.07~(25\%)}$ & 0.13 \\
 \hline
 \multirow{6}{4em}{COMPAS} & Precision & $\bm{0.67 \pm 0.04}$ & ${0.60 \pm 0.09~(10.45\%)}$  & 0.52\\
 &Unfairness ($\mathds U$)& ${0.14 \pm 0.04}$ & $\bm{0.09 \pm 0.06~(35.71\%)}$  & 0.13\\
 \cline{2-5}
 & Precision & $\bm{0.57 \pm 0.07}$ & ${0.53 \pm 0.04~(7.02\%)}$ & 0.51 \\ 
 &Unfairness ($\mathds U$)& $ {0.07 \pm 0.07}$ & $\bm{0.05 \pm 0.05~(28.57\%)}$ & 0.12 \\
 \cline{2-5}
 & Precision & $\bm{0.59 \pm 0.05}$ & ${0.58 \pm 0.06~(1.69\%)}$ & 0.53 \\ 
 &Unfairness ($\mathds U$)& $ {0.09 \pm 0.05}$ & $\bm{0.06 \pm 0.03~(33.33\%)}$ & 0.16 \\
 \hline
\end{tabular}
}
    \caption{Out-of-sample testing results (mean ± standard deviation). The numbers inside the parentheses represent the average reduction compared to the No-Fair model.}
    \label{tab:real_data}
\end{table} 
We generate the selection process three times following the aforementioned procedure for each dataset. We compare the performance of two different models: the No-Fair model, where the unfairness control parameter is set to  $\eta=1$ in~\eqref{eq:refor-probtrust_multi}, and the Fair model, where the unfairness control parameter $\eta$ is set to the respective empirical unfairness score. 

We conduct out-of-sample experiments with a training dataset of size $N=200$. The selection ratios are set to $\overline{\alpha}_1 = 0.7, \overline{\alpha}_2 = 0.35$ and $\underline{\alpha}_2 = 0.2$. The results of all experiments are averaged over 20 random trials. Table~\ref{tab:real_data} demonstrates the superior performance of $\epsilon$-IPWMFS. As we can see, both the No-Fair and Fair models achieve high precision compared with the existing selection policy. Besides, the Fair model can achieve much lower unfairness scores with a negligible decrease in precision compared with the No-Fair model.

\section*{Acknowledgments}
Z.~Jia and G.~Hanasusanto are funded in part by the National Science Foundation under grants 2342505 and 2343869. P.~Vayanos is funded in part by the National Science Foundation under grant 2046230. W.~Xie is funded in part by the National Science Foundation under grants 2246414 and 2246417. They thank the four anonymous referees whose reviews helped substantially improve the quality of the paper.

\appendix

\bibliography{aaai24}

\newpage

\section*{Appendix}
\section{Proofs in the main text}
In this section, we provide the proofs for Proposition~\ref{prop:mpm_conservative_multi} and Theorem~\ref{thm:refor-probtrust_multi} in Section~\ref{sec:mip_formulation}.

\subsection{Proof of Proposition~\ref{prop:mpm_conservative_multi}}\label{subsec:proof2}

\begin{proof}
Consider any feasible solution $( {\bm W^{[t]}}, b_t )_{t=1}^T$ to~\eqref{eq:appromodel_multi} and any distribution $\hat{\PP}^{\text{IPW}}$. We first prove the feasibility of the fairness constraint in~\eqref{eq:exactmodel_multi}. For any $\epsilon \in \Real_{++}$, we have
 \begin{align*}
        & \mathds U( \mcal C_T(\bm X^{[T]}), \hat{\PP}^{\text{IPW}} ) \\
         =&\max \left\{ \begin{array}{l}
        \hat{\PP}^{\text{IPW}}_{01} ( \bm W^{[T]} \cdot \bm X^{[T]} + b_T > 0 ) \\
        \;- \hat{\PP}^{\text{IPW}}_{11} ( \bm W^{[T]} \cdot \bm X^{[T]} + b_T > 0 ), \\
        \hat{\PP}^{\text{IPW}}_{11} ( \bm W^{[T]} \cdot \bm X^{[T]} + b_T > 0 ) \\
        \; - \hat{\PP}^{\text{IPW}}_{01} ( \bm W^{[T]} \cdot \bm X^{[T]} + b_T > 0 )
        \end{array}
        \right\}\\
        \leq &  \max \left\{ \begin{array}{l}
    \hat{\PP}^{\text{IPW}}_{01} ( \bm W^{[T]} \cdot \bm X^{[T]} + b_T > 0 ) \\
    - \hat{\PP}^{\text{IPW}}_{11} ( \bm W^{[T]} \cdot \bm X^{[T]} + b_T  \geq \epsilon ), \\
    \hat{\PP}^{\text{IPW}}_{11} ( \bm W^{[T]} \cdot \bm X^{[T]} + b_T > 0 ) \\
    - \hat{\PP}^{\text{IPW}}_{01} (\bm W^{[T]} \cdot \bm X^{[T]} + b_T  \geq \epsilon ) 
    \end{array}
    \right\} \\
         = & \mathds U_\epsilon (\mcal C_T( \bm X^{[T]}), \hat \PP^{\text{IPW}}) \leq \eta.
    \end{align*}
Therefore, $( {\bm W^{[t]}}, b_t )_{t=1}^T$ are also feasible to the fairness constraint in~\eqref{eq:exactmodel_multi}.
Next, by definition, we have 
    \begin{align*}
        \mcal C_{t+1}(\bm X^{[t+1]}) & \leq C_{t}^\epsilon(\bm X^{[t]}) \\
        & = \mathds{1}(  \bm W^{[t]} \cdot \bm X^{[t]}+ b_t \geq \epsilon ) \\
        & \leq \mathds{1}(  \bm W^{[t]} \cdot \bm X^{[t]}+ b_t > 0 )=  C_{t}(\bm X^{[t]}). 
    \end{align*}
This proves the feasibility of the penultimate constraint to~\eqref{eq:exactmodel_multi}. Lastly, we have
    \begin{align*}
         & \hat \PP^{\text{IPW}} ( \bm W^{[T]} \cdot \bm X^{[T]} + b_T  \leq 0 ) \\
        \leq & \hat \PP^{\text{IPW}} ( \bm W^{[T]} \cdot \bm X^{[T]} + b_T < \epsilon ) \\
        \leq & 1 - \underline{\alpha}_T,
    \end{align*}
    which proves the feasibility of the selection ratio requirement to~\eqref{eq:exactmodel_multi}. Consequently, the feasible set of problem \eqref{eq:appromodel_multi} is an \textit{inner} approximation of the feasible set of problem \eqref{eq:exactmodel_multi} for any $\epsilon \in \Real_{++}$. Therefore, an optimal solution $( {\bm W^{[t]}}^*, b_t^* )_{t=1}^T$ of problem \eqref{eq:appromodel_multi} is also feasible for problem \eqref{eq:exactmodel_multi}. By plugging in the optimal solution $( {\bm W^{[t]}}^*, b_t^* )_{t=1}^T$ to  \eqref{eq:exactmodel_multi}, we have $f^\star_{opt} \ge f^\star$. This completes the proof.
\end{proof}

\subsection{Proof of Theorem~\ref{thm:refor-probtrust_multi}}\label{subsec:proof3}

\begin{proof}
We start with the reformulation for the objective function of~\eqref{eq:appromodel_multi}. By definition of conditional probability, we can rewrite the precision as 
\begin{align*}
      \hat{\PP}^{\text{IPW}} ( Y = 1~|~ \mcal C_T (\bm X^{[T]})=1  ) \\
 = \frac{ \hat{\PP}^{\text{IPW}} ( Y = 1, \mcal C_T(\bm X^{[T]}) = 1 ) }{\hat{\PP}^{\text{IPW}} (\mcal C_T(\bm X^{[T]}) = 1 )}.
    \end{align*}
Using the fact that the final stage selection constraint requires $0 < \underline{\alpha}_T \leq \hat \PP^{\text{IPW}} (\mcal C_T(\bm X^{[T]}) = 1 )$, instead of maximizing the precision, we can minimize the reciprocal of it. By introducing an epigraphical variable $f$, which we minimize, and moving the reciprocal of the precision into the constraint, we have
   \begin{align*}
       & 
 \frac{ \hat{\PP}^{\text{IPW}} (\mcal C_T(\bm X^{[T]}) = 1 ) }{ \hat{\PP}^{\text{IPW}} ( Y = 1, \mcal C_T(\bm X^{[T]}) = 1 )} \\
 = & \renewcommand*{\arraystretch}{1.2}
           \left\{  \begin{array}{cll}
                \min &  f \\
                \textrm{s.t.} & f \in \Real \\
                 & 1 \leq f  \\
                 &  \hat{\PP}^{\text{IPW}} (\mcal C_T(\bm X^{[T]}) = 1 ) \\
                 & \quad \leq f \hat{\PP}^{\text{IPW}} ( Y = 1, \mcal C_T(\bm X^{[T]}) = 1),  \\
            \end{array} \right. 
    \end{align*}
where the first constraint comes from the fact that the reciprocal of a conditional probability is at least one. For any $t \in [T]$ and $i \in I^T$, we define 
\begin{align*} 
     g_{ti} &= \mathds{1}(C_t(\hat \x^{[t]}_i) = 1) = \mathds{1} (\bm W^{[t]} \cdot \hat \x_i^{[t]}+ b_t > 0)   \\
     p_{ti} &= \mathds{1}(C_t^\epsilon(\hat \x^{[t]}_i) = 1) = \mathds{1} (\bm W^{[t]} \cdot \hat \x_i^{[t]}+ b_t < \epsilon) . 
\end{align*}
Using big-M parameter $M$, we can reformulate the above indicator function as
\begin{align*} 
\left\{
       \renewcommand*{\arraystretch}{1.2}
            \begin{array}{cl}
                 & \bm g_t \in \{0, 1\}^{|I^T|},~\bm p_t \in \{0, 1\}^{|I^T|} \\
                 & -M(1-g_{ti}) \leq \bm W^{[t]} \cdot \hat \x_i^{[t]}+ b_t \leq M g_{ti}  \\
                 & \epsilon  -  \bm W^{[t]} \cdot \hat \x_i^{[t]} - b_t \leq M p_{ti}.
            \end{array}
        \right. 
\end{align*}
Recalling the definition of $\hat{\PP}^{\text{IPW}}$,
\begin{align*}
    \hat \PP^{\text{IPW}} = \sum_{i=1}^{|I^t|} \frac{ \beta_i^t }{\mathbf{e}^\intercal \bm \beta^t} \delta_{(\hat \x_i^{[t]}, \hat a_i, \hat y_i) }.
\end{align*} 
we can thus rewrite the above minimization problem as
\begin{align*}
    & \renewcommand*{\arraystretch}{1.2}
           \left\{  \begin{array}{cll}
                \min &  f \\
                \textrm{s.t.} & f \in \Real,~\bm g_T \in \{0,1\}^{|I^T|} \ \\
                 & 1 \leq f  \\
                 & \sum_{i=1}^{|I^T|} \beta_i^T (g_{Ti})^2 \leq f  \sum_{i \in \mcal I_1 }\beta_i^T g_{Ti}, \\
            \end{array} \right.
    \end{align*}
where $\mcal I_1 = \{i \in I^T:  \wh y_i = 1 \}$, $M$ is the big-M parameter. Next, we provide the reformulation for the upper bound selection ratio requirement constraint of~\eqref{eq:appromodel_multi} at $t \in [T]$. By definition, we have
    \begin{align*}
       & \hat{\PP}^{\text{IPW}} ( \bm W^{[t]} \cdot \bm X^{[t]} + b_t > 0 ) \leq \overline{\alpha}_t \\
       \Longleftrightarrow & \frac{ \sum_{i=1}^{|I^T|} \beta_i^t \mathds{1}( \bm W^{[t]} \cdot \hat \x_i^{[t]}  + b_t > 0 )  }{\mathbf{e}^\intercal \bm \beta^t} \leq \overline{\alpha}_t \\
       \Longleftrightarrow & \left\{
       \renewcommand*{\arraystretch}{1.2}
            \begin{array}{cl}
                 & \bm g_t \in \{0, 1\}^{|I^T|} \\
                 & \bm g_t^\intercal \bm \beta^t   \leq \overline{\alpha}_t(\mathbf{e}^\intercal \bm \beta^t) .
            \end{array}
        \right. 
\end{align*}
For the lower bound selection ratio requirements at the terminal stage, we have 
\begin{align*}
& \hat{\PP}^{\text{IPW}} ( \bm W^{[T]} \cdot \bm X^{[T]} + b_T < \epsilon ) \leq 1 - \underline{\alpha}_T \\
= & \left\{
       \renewcommand*{\arraystretch}{1.2}
            \begin{array}{cl}
                 & \bm p_T \in \{0, 1\}^{|I^T|} \\
                 & \bm p_T^\intercal \bm \beta^T  \leq (1 - \underline{\alpha}_T)(\mathbf{e}^\intercal \bm \beta^T).
            \end{array}
        \right.
\end{align*}
The admission requirement constraint of~\eqref{eq:appromodel_multi} can be written as 
\begin{align*}
       & \mcal C_{t+1}(\bm X^{[t+1]}) \leq \mcal C_{t}^\epsilon(\bm X^{[t]})  \\ 
       \Longleftrightarrow &
     \mathds{1}( \bm W^{[t+1]} \cdot \hat \x_i^{[t+1]} + b_{t+1} > 0)  \\
     & \hspace{1in}\leq \mathds{1}( \bm W^{[t]} \cdot \hat \x_i^{[t]} + b_{t} \geq \epsilon ) \\
     \Longleftrightarrow &  
     \mathds{1} (\bm W^{[t+1]} \cdot \hat \x_i^{[t+1]} + b_{t+1} > 0 ) \\
     & \hspace{1in} + \mathds{1}(\bm W^{[t]} \cdot \hat \x_i^{[t]} + b_{t} < \epsilon ) \leq 1.
    \end{align*}
    which can be reformulated as
   \begin{align*}
    \left\{ \renewcommand*{\arraystretch}{1.2}
                \begin{array}{cll}
                    & \bm g_{t+1} \in \{0, 1\}^{|I^T|},\; \bm p_t \in \{0, 1\}^{|I^T|} \\
                    & \bm g_{t+1}  + \bm p_{t} \leq \mathbf{e} .
               \end{array}
            \right.
    \end{align*}
Finally, we provide the reformulation for the fairness constraint of~\eqref{eq:appromodel_multi}. Define the index sets
    $\mcal I_{a1} = \left\{ i \in [N]: \wh a_i = a, \wh y_i = 1\right\}~ \forall a \in \mcal A$. For any fixed pair $(a, a') \in \{(0, 1), (1, 0)\}$, we have
\begin{IEEEeqnarray*}{rCl} 
&&\hat{\PP}_{a1}^{\text{IPW}} ( \bm W^{[T]}\cdot \bm X^{[T]} + b_T > 0 ) \\
&& \; - \hat{\PP}_{a'1}^{\text{IPW}} ( \bm W^{[T]}\cdot \bm X^{[T]} + b_T \geq \epsilon ) \\
&=&\hat{\PP}_{a1}^{\text{IPW}} ( \bm W^{[T]}\cdot \bm X^{[T]} + b_T > 0 ) \\
&& \; + \hat{\PP}_{a'1}^{\text{IPW}} ( \bm W^{[T]}\cdot \bm X^{[T]} + b_T < \epsilon ) - 1\\
&=&  \EE_{ \hat {\PP}^{\text{IPW}}}  \left[\wh p_{a1}^{-1} \mathds{1} ( \bm W^{[T]}\cdot \bm X^{[T]} + b_T > 0 )  \mathds{1}( (A, Y)=(a, 1) )\right.\\
&& \; \left. + \wh p_{a'1}^{-1} \mathds{1} ( \bm W^{[T]}\cdot \bm X^{[T]} + b_T < \epsilon ) \mathds{1} ( (A, Y) = (a', 1) ) \right] - 1,
\end{IEEEeqnarray*}
where $ \wh p_{a1} = \hat {\PP}^{\text{IPW}}(A = a, Y = 1)$ and  $ \wh p_{a'1} = \hat {\PP}^{\text{IPW}}(A = a', Y = 1)$. Using the definition of $\hat {\PP}^{\text{IPW}}$, we can rewrite the about expression as 
\begin{IEEEeqnarray*}{rCl} 
\left\{  \renewcommand*{\arraystretch}{1.2}
            \begin{array}{cl}
                & \bm g_T \in \{0, 1\}^{|I^T|},~ \bm p_T \in \{0, 1\}^{|I^T|}\\
                & \displaystyle \frac{1}{  \sum_{i \in \mcal I_{a1}} \beta_i^T } \sum_{i \in \mathcal{I}_{a1}}\beta_i^T g_{Ti} + \frac{1}{ \sum_{i \in \mcal I_{a'1}} \beta_i^T } \sum_{i \in \mathcal{I}_{a'1}} \beta_i^T p_{Ti} - 1 .
           \end{array}
        \right.
\end{IEEEeqnarray*}
Combining all the above analysis, we have the result in Theorem~\ref{thm:refor-probtrust_multi}.
\end{proof}

\section{Real Datasets}
In this section, we provide the details for the datasets in Section~\ref{sec:numerical}.

\begin{table}[htb!]
    \centering
    \resizebox{1\columnwidth}{!}{%
    \begin{tabular}{||p{2.5cm}|c|c|c||}
    \hline
         \textbf{Dataset}& \textbf{Covariates $d$} & \textbf{Protected Attribute} & \textbf{Number of Samples}  \\
         \hline\hline
         Adult  & 12 & Gender & 32561, 12661 \\
         German & 19 &  Gender & 1000\\ 
         COMPAS & 10 & Ethnicity & 6172 \\
         \hline
     \hline
    \end{tabular}}
    \caption{Summary of Datasets.}
    \label{tab:statistics}
\end{table}
 The Adult dataset, also known as ``Census Income'' dataset, is taken from a 1994 Census database. The goal is to determine whether a person’s annual income exceeds \$50,000 or not. It contains 13 features concerning demographic characteristics of 45,222 instances, and we consider gender as the sensitive attribute. 
 
 The German Credit Risks dataset consists of samples of bank account holders with good or bad credit. The data are collected from 1,000 individuals, and we take gender as the sensitive attribute.

The COMPAS dataset consists of the variables of a commercial algorithm called COMPAS (Correctional Offender Management Profiling for Alternative Sanctions), used to predict a convicted criminal’s likelihood of recidivism within two years. The data contains 6,172 data points, and we use ethnicity as the sensitive attribute. 

\end{document}